\documentclass[conference]{IEEEtran}
\usepackage{times}
\usepackage{graphicx}
\usepackage[numbers]{natbib}
\usepackage{multicol}
\usepackage[bookmarks=true]{hyperref}
\usepackage{amsfonts}
\usepackage{stfloats}
\usepackage{placeins} 
\usepackage{cuted}
\usepackage{caption}
\usepackage{amsmath}
\usepackage{booktabs}
\usepackage{multirow}
\usepackage{graphicx}
\usepackage{hyperref}
\usepackage{xcolor}
\usepackage{todonotes}
\definecolor{sharedpurple}{RGB}{141,59,255} 
\definecolor{sharedred}{RGB}{201, 79, 79} 
\definecolor{sharedblue}{RGB}{110, 212, 249} 
\definecolor{sharedyellow}{RGB}{237,161,0} 
\newcommand{\accro}[0]{TactAlign}

\begin{document}


\title{\accro: Human-to-Robot Policy Transfer via Tactile Alignment}

\author{Author Names Omitted for Anonymous Review. Paper-ID [534]}

\author{
\authorblockN{
Youngsun Wi\textsuperscript{1},
Jessica Yin\textsuperscript{2},
Elvis Xiang\textsuperscript{1},
Akash Sharma\textsuperscript{3},
Jitendra Malik\textsuperscript{4},
Mustafa Mukadam\textsuperscript{5}, \\
Nima Fazeli\textsuperscript{1,*},
Tess Hellebrekers\textsuperscript{6,*}
}
\authorblockA{
\textsuperscript{1}University of Michigan 
\textsuperscript{2}Nvidia 
\textsuperscript{3}Amazon Frontier AI \& Robotics 
\textsuperscript{4}UC Berkeley
\textsuperscript{5}University of Washington 
\textsuperscript{6}Microsoft Research
}
\small \textsuperscript{$\dagger$}This work was partially done during Youngsun Wi's Meta FAIR internship. \textsuperscript{*}Equal advising. \\
}


%

\maketitle
\vspace{-1.0em}
\begin{strip}
\centering
\includegraphics[width=\textwidth]{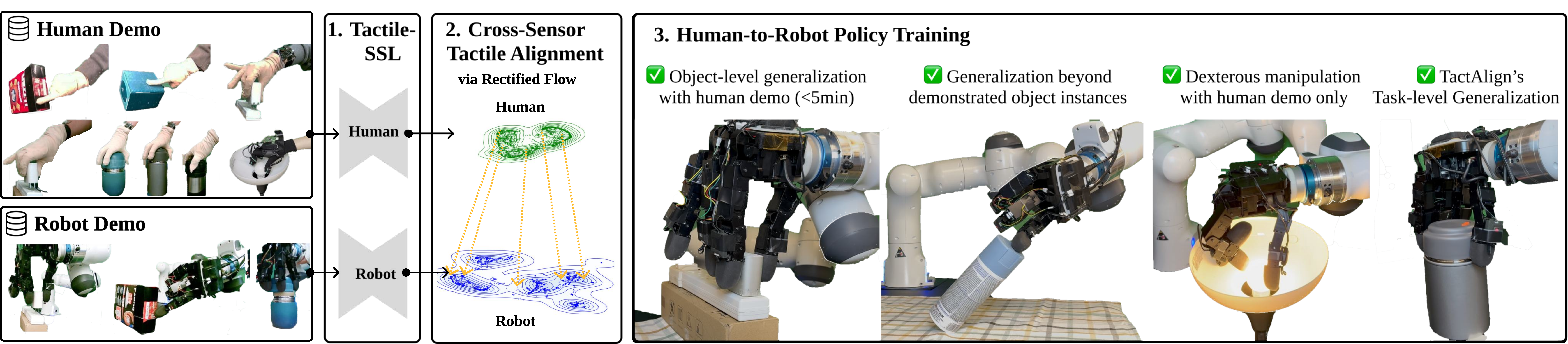}
\captionof{figure}{\small We propose \accro, a cross-sensor tactile alignment method for cross-embodiment human-to-robot policy transfer. Given unpaired human (tactile glove) and robot demonstrations, \accro\ uses a rectified flow to map glove tactile features into the robot tactile space. This alignment enables effective tactile policy co-training on pivoting, insertion, and lid closing tasks. With only a few minutes of human demonstrations, the resulting policies generalize to unseen objects instances. Importantly, the same learned alignment can be reused to train policies on unseen tasks. We also demonstrate zero-shot human-to-robot dexterous manipulation on a light bulb screwing task.}
\label{fig:teaser}
\end{strip}

\begin{abstract}
Human demonstrations collected by wearable devices (e.g., tactile gloves) provide fast and dexterous supervision for policy learning, and are guided by rich, natural tactile feedback. However, a key challenge is how to transfer human-collected tactile signals to robots despite the differences in sensing modalities and embodiment. Existing human-to-robot (H2R) approaches that incorporate touch often assume identical tactile sensors, require paired data, and involve little to no embodiment gap between human demonstrator and the robots, limiting scalability and generality. We propose \accro, a cross-embodiment tactile alignment method that transfers human-collected tactile signals to a robot with different embodiment. \accro\ transforms human and robot tactile observations into a shared latent representation using a rectified flow, without paired datasets, manual labels, or privileged information. Our method enables low-cost latent transport guided by hand-object interaction-derived pseudo-pairs. We demonstrate that \accro\ improves H2R policy transfer across multiple contact-rich tasks (pivoting, insertion, lid closing), generalizes to unseen objects and tasks with human data ($\leq 5$ minutes), and enables zero-shot H2R transfer on a highly dexterous tasks (light bulb screwing). \\
\vspace{0.25em}\noindent
\textsc{Website:} \href{https://yswi.github.io/tactalign/}{\texttt{yswi.github.io/tactalign/}}
\end{abstract}

\IEEEpeerreviewmaketitle

\section{Introduction}

As scaling data becomes increasingly important in robot learning, human demonstrations have emerged as a compelling data source:  human trajectories can be collected 2-3x faster than robot teleoperation data~\cite{punamiya2025egobridge}, while also being inherently dexterous and multi-modal. 
Critically, our manipulation skills are guided by our rich, multi-sensory tactile feedback. However, most existing human-to-robot (H2R) approaches omit tactile feedback entirely and instead focus on transferring more readily available observations such as egocentric vision or state–action pairs in configuration space. As a result, tactile feedback, despite its central role in dexterous and contact-rich manipulation, has remained largely underexplored in H2R learning. This raises a key question: \textit{how can human touch, collected through through wearable tactile devices, be effectively represented and transferred to robots?}

Recent work has begun to address this question by incorporating tactile sensing into human demonstrations for robot policy learning ~\cite{yin2025osmo, yu2025mimictouch, adeniji2025feel, xudexumi, fang2025dexop, zhang2025unitachand, wang2023mimicplay}. While effective, many of these approaches assume identical tactile sensors or little to no embodiment gap, which simplifies transfer but limits applicability across diverse robot hands. A concurrent work, UniTacHand~\cite{zhang2025unitachand}, addresses cross-sensor tactile transfer, but relies on strict spatiotemporal correspondence between human and robot throughout task demonstrations. This strict pairing can be prohibitively difficult to maintain during contact-rich interactions involving sliding contact or dynamic object motion necessary for general manipulation. In contrast, our proposed method enables tactile transfer from unpaired datasets of the same task without requiring such pairing assumptions.

Beyond H2R settings, cross-sensor tactile transfer has also been studied. However, existing methods primarily focus on static contact scenarios~\cite{rodriguez2024contrastive, rodriguez2024touch2touch, grella2025touch, feng2025anytouch} and coarse categorical alignment objectives~\cite{zhao2025transferable, feng2025anytouch}, leaving their effectiveness for continuous tactile reasoning during dynamic interactions such as sliding contact or object motion unclear. Moreover, these approaches often rely on paired supervision or labels, limiting scalability across heterogeneous sensors and robots.

We present \accro, a method to transfer human tactile observations collected with a wearable tactile device (e.g., a tactile glove~\cite{yin2025osmo}) to robots with heterogeneous tactile sensors. The objective of this cross-sensor tactile transfer is to enable effective cross-embodiment human-to-robot policy transfer. At a high level, \accro\ performs tactile alignment in two stages. First, it pretrains human and robot tactile encoders independently using self-supervised learning to obtain modality-specific latent representations. Second, it learns a cross-sensor alignment via rectified flow using pseudo-pairs derived from hand-object interactions, without requiring explicitly paired datasets. Rectified flow is well suited to learning mappings under noisy pseudo-pairs \cite{liuflow} arising from the non-unique relationship between hand-object motion and tactile observations. 
Together, these design choices simplify data requirements, allowing human demonstrations to be effectively leveraged for robot learning despite heterogeneous embodiment.

The core contributions of our work are:
\begin{itemize}
    \item We propose \accro, a method for aligning cross-sensor tactile data from \textit{unpaired} demonstrations of the same task. TactAlign leverages rectified flow with noisy pseudo-pairs to learn a latent mapping that enables H2R policy transfer  between humans and robots equipped with heterogeneous tactile sensors. 
    \item We show \accro\ improves H2R co-training success by $+59$\% (vs. no tactile) and $+51$\% (vs. no alignment), and generalizes to human-only objects ($+59$\%, vs. robot-only) and unseen objects ($+54$\%, vs. robot-only)  using $\leq$5 minutes of human data across contact-rich tasks: pivoting, insertion, lid closing.
    \item Finally, we show that \accro\ enables zero-shot dexterous policy transfer from human data for light-bulb screwing, achieving a $+100$\% improvement over policies trained without tactile input or alignment.
\end{itemize}

\section{Related Works}
\subsection{Human-to-Robot Transfer}

Prior work has shown leveraging fast and diverse human demonstrations enables scalable and generalizable robot policies. Most existing human-to-robot transfer methods operate in visual \cite{wang2023mimicplay, kim2025uniskill, punamiya2025egobridge, kareer2025emergencehumanrobottransfer, bharadhwaj2024gen2act, bharadhwaj2023zero} or kinematic spaces~\cite{tao2025dexwild,  kareeregomimic, liu2025egozero, chen2025flowing, guzey2025dexterity, lepert2025masquerade}. More recently, a small number of works have begun incorporating tactile sensing into this paradigm, transferring human tactile signals to robots either via simplified grippers~\cite{yu2025mimictouch, adeniji2025feel} or through wearable systems tailored to specific robot hands~\cite{xudexumi, fang2025dexop, yin2025osmo}. These methods highlight the value of human tactile data by enabling demonstrators to feel and react to touch and transferring human tactile measurements to robot policy learning, but they assume humans and robots share the same tactile sensors. 
A concurrent work, UniTacHand~\cite{zhang2025unitachand}, addresses cross-sensor tactile transfer, but relies on spatially and temporally strictly paired human-robot data. In contrast, TactAlign enables tactile transfer from unpaired datasets of the same task without requiring such strict correspondences.

\subsection{Wearable Devices for Data-Collection}
Recent works have explored wearable devices for more intuitive demonstration capture, particularly for dexterous robot hands. Compared to teleoperation, wearables allow the human demonstrator to interact directly with real objects with natural dexterity and haptic feedback, rather than controlling a different robot embodiment through intermediate devices and kinematic mismatches. However, current wearable devices necessitate a trade-off between the demonstrator's degrees of freedom (DOF) and tactile signal richness. While handheld and fingertip-based approaches \cite{choi2026wild, yu2025mimictouch, adeniji2025feel} achieve high-fidelity 3D force sensing, they restrict the user to low-DOF parallel-jaw grasps. To support multi-fingered manipulation, exoskeleton devices \cite{xudexumi, fang2025dexop, zhang2025doglove} mechanically constrain human movement to match robot linkages, improving retargeting accuracy but sacrificing the dexterity of the human hand. Flexible gloves preserve the full DOFs of human dexterity but can be limited to kinematics-only data \cite{tao2025dexwild,yin2025dexteritygen} or normal forces only \cite{luo2024tactile,SSundaram:2019:STAG}. In this work, we use the OSMO tactile glove \cite{yin2025osmo}, which combines the high dexterity of flexible gloves with rich shear and normal force sensing.

\subsection{Cross-Sensor Tactile Alignment}
A central challenge in tactile learning is handling heterogeneous sensor modalities while enabling effective knowledge transfer across them. Some prior works address this by learning shared representations across vision-based tactile sensors~\cite{higuera2024sparsh}, but these approaches do not provide explicit one-to-one alignment or support direct cross-modal transfer. Other methods pursue explicit transfer through paired supervision or shared intermediate representations ~\cite{rodriguez2024contrastive, grella2025touch, feng2025anytouch, zhao2025transferable, rodriguez2024touch2touch, chen2026genforce}. However, these approaches either emphasize geometric aspects of touch, which limits their ability for dexterous manipulation that involves shear or sliding contact \cite{rodriguez2024touch2touch, rodriguez2024contrastive, grella2025touch}, focus on coarse categorical alignment \cite{feng2025anytouch}, or rely on paired data collected using hand-designed 3D-printed objects~\cite{chen2026genforce, rodriguez2024touch2touch}. In contrast, our approach enables dense tactile alignment from natural task demonstrations, including sliding and dynamic motion, without requiring explicit labels or strictly paired data.


\begin{figure*}
    \centering
    \includegraphics[width=1\linewidth]{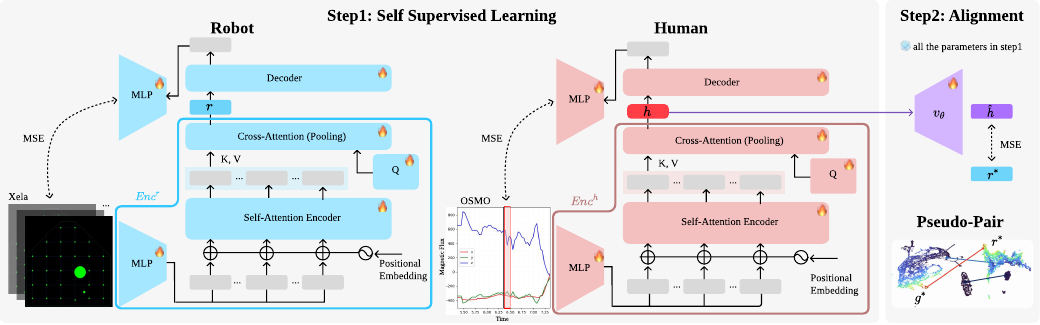}
    \caption{\small \textbf{Tactile Alignment Overview. }Our method consists of two stages: self-supervised representation learning and cross-embodiment alignment via pseudo-pairs. We use a learnable length-1 query between the encoder and decoder to produce a fixed-dimensional latent representation via cross-attention pooling. A learnable length 1 query is implemented between the encoder and decoder to output a fixed-dimensional latent representations after the cross-attention module. In step2, we aggregate the learned latents from both domains to construct pseudo-pairs $(h^*, r^*)$, and learn a velocity field \textcolor{sharedpurple}{$v_\theta$} that transports the glove latent distribution to the robot latent distribution.}
    \label{fig: architecture}
\end{figure*}

\section{Methodology}

\subsection{Problem Statement}
Our goal is to learn a latent-space mapping that transfers human tactile observations
to robot tactile observations.
We assume access to two offline datasets of task demonstration trajectories:
one collected by a human wearing a tactile glove, denoted by $H$, and another
collected from a dexterous robot with heterogeneous fingertip tactile sensors,
denoted by $R$. The robot dataset is smaller in scale due to the higher cost of robot data collection. The datasets consist of demonstration trajectories,
$H=\{\mathcal{T}^h_1,\ldots,\mathcal{T}^h_N\}$ and
$R=\{\mathcal{T}^r_1,\ldots,\mathcal{T}^r_M\}$.
Each human trajectory is represented as
$
\mathcal{T}^h=\{(F^h_1,P^h_1,w^h_1),\ldots,(F^h_l,P^h_l,w^h_l)\},
$
while each robot trajectory is
$
\mathcal{T}^r=\{(F^r_1,P^r_1,w^r_1),\ldots,(F^r_l,P^r_l,w^r_l)\},
$
where the length $l$ varies across trajectories and $t$ denotes the time step. At each time step $t$, tactile observations and poses from all $K$ fingertips are represented as 
$
F_t=(f_{t,1},\ldots,f_{t,K}), \quad
P_t=(p_{t,1},\ldots,p_{t,K}),
$
where $f_{t,k}$ and $p_{t,k}$ denote the tactile observation and pose of fingertip $k$,
respectively, and $w_t$ denotes the wrist pose. Superscripts $h$ and $r$ indicate human and robot.

We denote latent tactile mapping as $g: T^h \rightarrow T^r$,
where $T^h, T^r \in \mathbb{R}^d$ are latent spaces derived from the human dataset
$H$ and the robot dataset $R$, respectively.
We assume that a subset of trajectories in $H$ and $R$, denoted by
$A^h \subset H$ and $A^r \subset R$, correspond to the same task and share
start and end object states. For these trajectories, we assume access to object pose estimates extracted from images, which are used only to establish initial correspondences. We represent these subsets as
$A^h=\{(F^h_1,P^h_1,w^h_1,o^h_1),\ldots,(F^h_l,P^h_l,w^h_l,o^h_l)\},$
$A^r=\{(F^r_1,P^r_1,w^r_1,o^r_1),\ldots,(F^r_l,P^r_l,w^r_l,o^r_l)\},$
where $o_t$ denotes the object pose at time $t$.

\subsection{Tactile Self-supervised Learning \label{sec: tactile-ssl}}

We represent per-fingertip tactile observations from two embodiments for the human glove and the robot. We use $f_i^g, p_i^g$ and $f_j^r, p_j^r$ to denote the tactile observation and pose
of a single fingertip at independent time indices $i$ and $j$ for the human and robot
trajectories, respectively. Here, the human tactile observation has size $w^h \times n^h \times d^h$, while the robot tactile signal is $w^r \times n^r \times d^r$, where $w^h$ and $w^r$ are a 0.1 second time window as in \cite{sharma2025self}, $n^h$ and $n^r$ correspond to the spatial resolution, $d^h$ and $d^r$ are dimensions for the glove and robot, respectively. To accommodate the heterogeneous sensing modalities, we learn unique encoders and decoders for the human and robot tactile signals in a self-supervised manner, using a mean squared error (MSE) reconstruction loss to preserve modality-specific structure. Our architecture (Fig.~\ref{fig: architecture} left) is based on JEPA \cite{assran2023self} with the decoder adapted from the online probe module in \cite{higuera2024sparsh, sharma2025self}. As a result, we get pretrained encoders for human and robot observations: 
\begin{align}
    \textcolor{sharedred}{Enc^h}(f^h_i) = h_i \in \mathbb{R}^{d}~~ \text{and} ~~\textcolor{sharedblue}{Enc^r}(f^r_i) = r_i \in \mathbb{R}^{d},
    \label{eq: tactile encoder}
\end{align}
where $\mathbb{R}^{d}$ is the shared latent tactile feature dimension. Since the two tactile signals differ in dimensionality, we implemented cross attention based pooling module at the end of each encoder for tactile features with consistent length \cite{santos2016attentive}.


 \begin{figure*}
    \centering
    \includegraphics[width=1\linewidth]{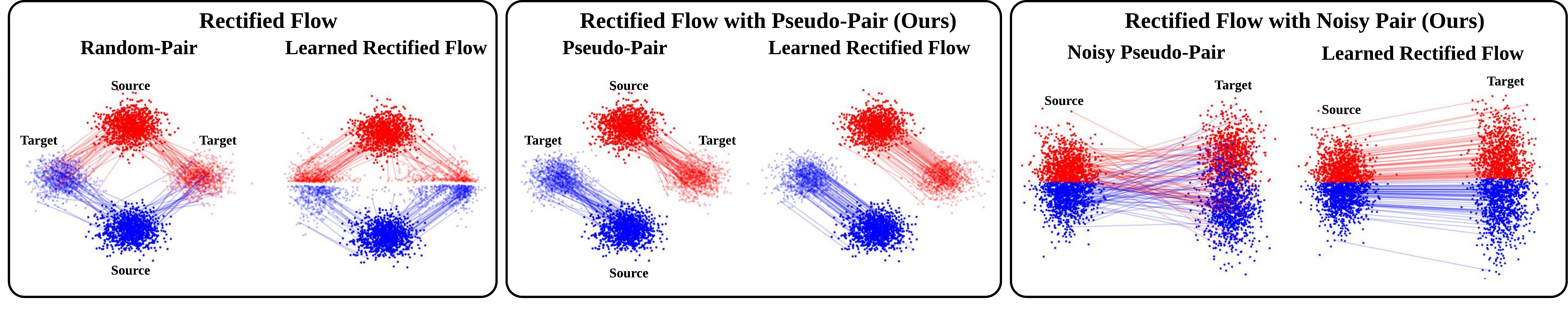}
    \caption{\small Red and blue indicate two subsets of the source distribution; training uses the provided pairs (lines), with colors preserved at $\alpha = 0.2$ for the target samples associated with each pair (left of each panel) and their transformed targets (right of each panel). \textbf{First:} Standard rectified flow \cite{liuflow} learns a low-cost transport between two distributions by training on randomly. \textbf{Second:} We propose using pseudo-pairs to the rectified flow for guiding the velocity field toward desired correspondences between the source and target distributions. \textbf{Third: }Despite noise in the pseudo-pairs, the learned rectified flow remains robust and converges to an efficient transport map between the two distributions.}
    \label{fig: rectified flow}
\end{figure*}


\subsection{Pseudo-Pair Extraction from Demonstrations \label{sec: pseudo-pair}}
We extract human hand and object pose from each trajectory in a single pass,
without requiring privileged information such as 3D object models
(Appendix~\ref{sec: data post processing}). Then, we construct initial cross-domain
correspondences from transitions:
\[
O^h_i = (p^h_i, o^h_i, p^h_{i+1}, o^h_{i+1}), \quad
O^r_j = (p^r_j, o^r_j, p^r_{j+1}, o^r_{j+1}).
\]
We define the similarity metric between transitions as
\begin{equation}
\begin{aligned}
S(O^h_i, O^r_j) =
&\lVert p^h_i - p^r_j \rVert
+ \lVert o^h_i - o^r_j \rVert \\
&+ \lambda \left\lVert
\hat{\Delta} p^h_i - \hat{\Delta} p^r_j
\right\rVert
+ \lambda \left\lVert
\hat{\Delta} o^h_i - \hat{\Delta} o^r_j
\right\rVert.
\end{aligned}
\end{equation}
In this section, we operate in a normalized pose space, where positions and orientations are rescaled using task-level statistics; we reuse $p$ and $o$ to denote the normalized values (Appendix~\ref{sec: appendix psuedo-pairs}). The scalar $\lambda$ balances pose and velocity terms, and a single value is used across all (Appendix~\ref{sec: balancing term}). 

Using this metric, we create a set of latent tactile pseudo pairs $P = \{(h_i^\ast, r_j^\ast) \mid S(O^h_i, O^r_j) < \delta\}$, where $\delta$ is a global similarity threshold. We construct the pseudo pairs from $A^h$ and $A^r$ between human-robot demonstrations from the same task, object, reset state and goal state. These pseudo pairs are inherently noisy and thus serve only as an initial alignment guide. We further refine $P$ using binary contact filtering. Specifically, if
$\lVert f^h_i \rVert < \delta_h$ or $\lVert f^r_j \rVert < \delta_r$,
the corresponding observations $O^h_i$ and $O^r_j$ are classified as non-contact; otherwise, they are treated as contact. We retain only pseudo pairs that map contact-to-contact and non-contact-to-non-contact states. This simple filtering is effective because contact transitions are often subtle in configuration space but are readily distinguishable in raw tactile observations.


\subsection{Tactile Alignment via Rectified Flow \label{sec: rectified flow}}

We formulate the cross-embodiment tactile alignment problem as a rectified flow \cite{liuflow} problem over a conditional distribution $p(x \mid t, z)$. The variable $x$ denotes a latent tactile state evolving over normalized time $t \in [0,1]$, while the conditioning variable $z = (h_i^\ast, r_j^\ast)$ is obtained from pseudo pairs constructed offline across different tasks. Unlike the original rectified flow \cite{liuflow}, which relies on random pairings, we guide the flow using pseudo-pairs extracted from offline hand-object interactions, providing coarse initial correspondences (Fig.~\ref{fig: rectified flow}).

Our goal is to learn a velocity field $v_\theta$ that transports human tactile features to the robot's (Fig.~\ref{fig: architecture} right). Specifically, for each pseudo pair $(h_i^\ast, r_j^\ast) \in P$, we define an interpolated latent state
$x_t = t \cdot h_i^\ast + (1 - t) r_j^\ast$ and constant velocity $h_i^\ast - r_j^\ast$ at $x_t$. 
The velocity field \textcolor{sharedpurple}{$v_\theta$} is trained by solving a least-squares regression problem over $t \in [0,1]$:
\begin{equation}
\min_{v_\theta} \sum_{(h_i^\ast, r_j^\ast) \in P} \int_0^1
\left\| (h_i^\ast - r_j^\ast) - \textcolor{sharedpurple}{v_\theta}(x_t, t) \right\|^2 dt . \label{eq: flow loss}
\end{equation} This process naturally performs latent ``rewiring", effectively handling crossings and transport cost reductions while learned from noisy pairs as in Fig.~\ref{fig: rectified flow}. During inference time, we simulate the resulting ODE, $dx_t = \textcolor{sharedpurple}{v_\theta}(x_t, t)$, and transform human tactile feature from $h_i$ to $\hat h_i$ as
\begin{align}
    g_\theta(h_i) = \hat h_i = \int^{1}_{0}\textcolor{sharedpurple}{v_\theta}(x_t, t) dt ~~~ \text{with}~ x_0 = h_i. \label{eq: glove tactile feature transform}
\end{align}
In practice, we solve Eq.~\ref{eq: glove tactile feature transform} via vanilla Euler method with constant step size \cite{liuflow}. Theoretically, the ODE can be solved either forward and backward as Eq.~\ref{eq: flow loss} is time-symmetric. Architecture and training details are in Appendix~\ref{sec: tactile alignment training details}.

\subsection{Human to Robot Policy Learning \label{sec: h2r policy}}
\begin{figure}[b!]
    \centering
    \includegraphics[width=0.8\linewidth]{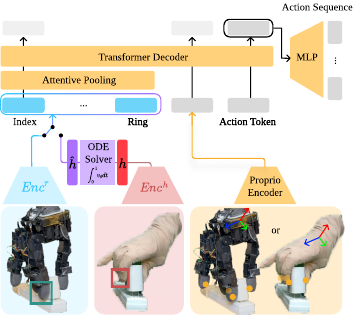}
    \caption{\small \textbf{H2R Action Policy.}
Given either human or robot inputs, the shared policy follows a color-coded structure, representing \textcolor{sharedblue}{robot}, \textcolor{sharedred}{human}, and \textcolor{sharedyellow}{shared} modules. Human glove latent features are passed into an ODE solver via a learned velocity field. The proprioceptive encoder takes fingertip locations in \textcolor{sharedyellow}{yellow dots} and wrist orientation. Only the \textcolor{sharedyellow}{yellow} modules are trained; all others are frozen.}
    \label{fig: h2r policy}
\end{figure}

We learn a H2R policy,
adapted from ACT~\cite{zhao2023learning}, which is shared across both human and robot
embodiments (Fig.~\ref{fig: h2r policy}).
The policy is defined as
$\textcolor{sharedyellow}{\pi_\phi}(\{\hat{h}_{t,k}\}_{k=1}^K, P^h_t, w^h_t)=a^h_t$
and
$\textcolor{sharedyellow}{\pi_\phi}(\{r_{t,k}\}_{k=1}^K, P^r_t, w^r_t)=a^r_t$. Here, $\{\hat{h}_{t,k}\}_{k=1}^K$ and $\{r_{t,k}\}_{k=1}^K$ denote per-fingertip tactile
latent features for the human and robot, respectively. Both outputs $a^h_t, a^r_t$ consist of action chunks specifying desired fingertip locations and wrist orientation with respect to the robot base frame. For human demonstrations, the wrist orientation is adjusted by a constant offset equal to the difference between the average human and robot wrist orientations over the entire trajectory~\cite{kareeregomimic}. During execution, the policy runs at 10-30~Hz (Appendix~\ref{sec: policy roll-out details}).

\section{Experiments and Results}

\subsection{Hardware}

In our experiments, humans wear the OSMO glove \cite{yin2025osmo}, which provides three-axis magnetic tactile signals at the fingertips. Robot demonstrations are collected using Xela sensors mounted on the Allegro Hand fingertips. While both sensors are magnetic-based, they differ in sensing mechanisms, as OSMO uses a particle-based magnetic skin whereas Xela employs discrete magnet-based sensing, resulting in differences in signal scale and characteristics as well as spatial resolution (OSMO: $1\times3$; Xela: $30\times3$). We use a Franka Emika Panda arm and a RealSense D455 camera, and employ an ATI Gamma force–torque (F/T) sensor only in Sec.~\ref{sec: force estimation}.

\subsection{Dataset}
We summarize the datasets used at each step below. Details on tactile signal post-processing are provided in Appendix~\ref{sec: signal post processing}.

\subsubsection{\textbf{Tactile self-supervised learning}} Both human and robot tactile encoders are trained using a combination of play data ($\approx$10 minutes) and an in-domain tactile alignment dataset.

\subsubsection{\textbf{Tactile alignment via rectified flow}} We use a total of 100 robot demonstrations via kinesthetic teaching and 200 human demonstrations collected from two contact-rich manipulation tasks: pivoting and insertion. Each task involves a single shared object, and we intentionally collect diverse human demonstrations. At this stage only, we focus on tasks in which the object pose changes relative to the hand or world during interaction. 

\subsubsection{\textbf{Human-robot policy co-training}} We chose three representative contact-rich tasks: pivoting, insertion, and lid-closing. For each task, we collect 140–160 human demonstrations, where 100 demonstrations ($\approx$ 30 minutes) are from the same object seen by the robot (``seen-by-both'' object), and 20 demonstrations are collected for each additional ``human-only'' object  ($\approx$ 5 minutes per object). For robot data, we collect 50 via kinesthetic using a single training object  ($\approx$ 60 minutes).

\subsubsection{\textbf{Dexterous policy learning with human-only}} Here, we collect human data only, consisting of 20 demonstrations of light-bulb screwing. We use Manus glove \cite{manus_glove} with OSMO tactile sensors~\cite{yin2025osmo} for robust hand pose estimation under visual occlusions from the lamp shade and light bulb. We record fingertip poses only, as the Manus glove does not provide wrist pose information. During data collection, human demonstrators screw a light bulb into a fixed lamp; the human wrist pose is randomized across demonstrations but remains static within each demonstration.

\subsubsection{\textbf{Force prediction evaluation}}  We collect a force-labeled dataset using an F/T sensor for both robot and human tactile observations (Appendix~\ref{sec: force data collection setup}). The dataset includes 1,472 robot force samples (24:1 train:test split) and 1,527 human force samples used only for evaluation. Robot force data are used exclusively for training, while human force measurements are reserved for testing (Appendix~\ref{sec: force estimation}).

\subsection{Learned Rectified Flow}

\begin{figure}
    \centering
    \includegraphics[width=1\linewidth]{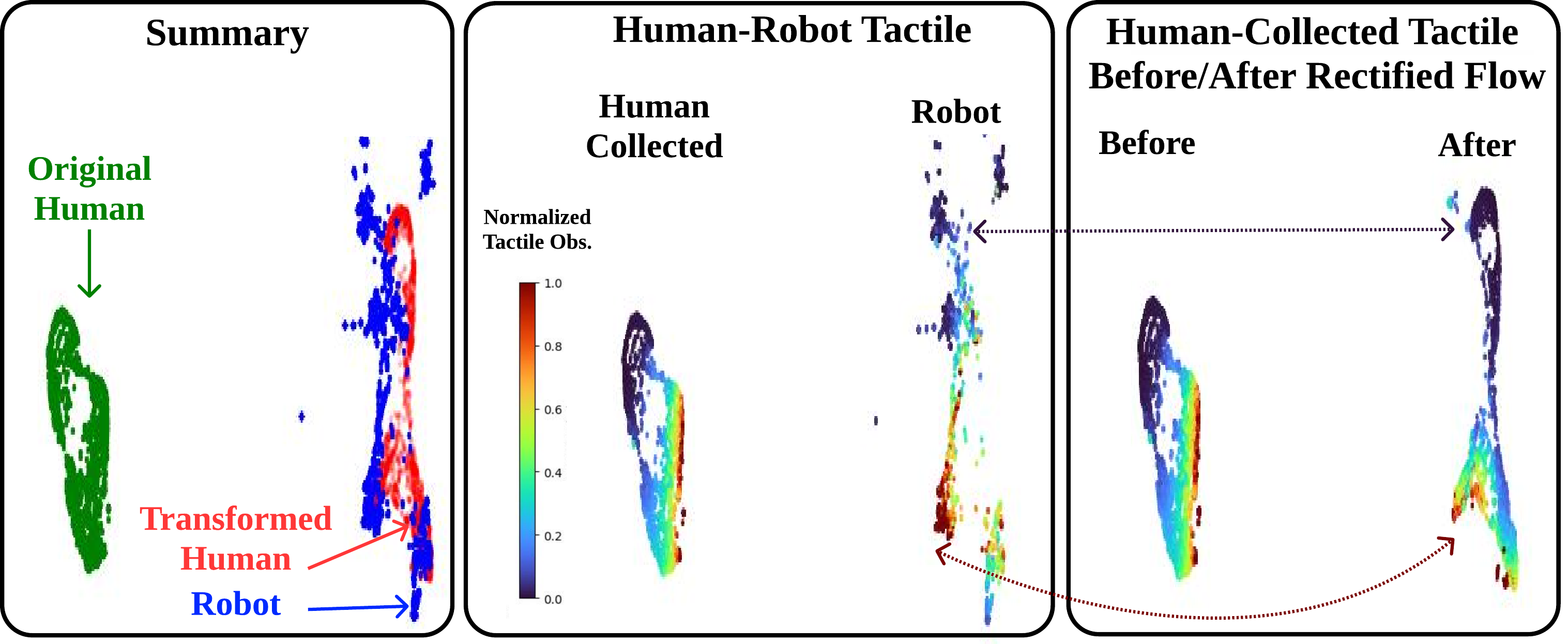}
    \caption{\small \textbf{Tactile Features UMAP Projections.} \textbf{First:} Rectified flow maps the glove latent distribution to overlap with the robot distribution. \textbf{Second \& Third:} Colors denote normalized raw tactile magnitude (0: no contact, 1: highest force/shear), computed separately for glove and robot data. As indicated by the arrows, the alignment exhibits a consistent cross-domain trend in contact force magnitudes, even though force is not used during training.}
    \label{fig: tactile latent distribution}
\end{figure}

We evaluate the learned human-to-robot rectified flow using UMAP projections~\cite{mcinnes2020umapuniformmanifoldapproximation} in Fig.~\ref{fig: tactile latent distribution} for both the pivoting and insertion tasks. After alignment, the human tactile features measured from the glove move toward the robot features and nearly overlap. Interestingly, we observe a consistent trend in normalized force magnitudes across domains: glove features associated with higher contact forces tend to map to robot features with similarly high normalized forces, and vice versa, despite force never being explicitly used during training. Quantitatively, the Earth Mover’s Distance (EMD)~\cite{kantorovich1960mathematical}, which measures distance between distributions while accounting for their geometry, between the human and robot tactile distributions decreases by 78\% after alignment, from 0.091 to 0.020. Further force-centric qualitative analysis is in Sec.~\ref{sec: force estimation}.

\begin{figure*}
    \centering
    \includegraphics[width=0.9\linewidth]{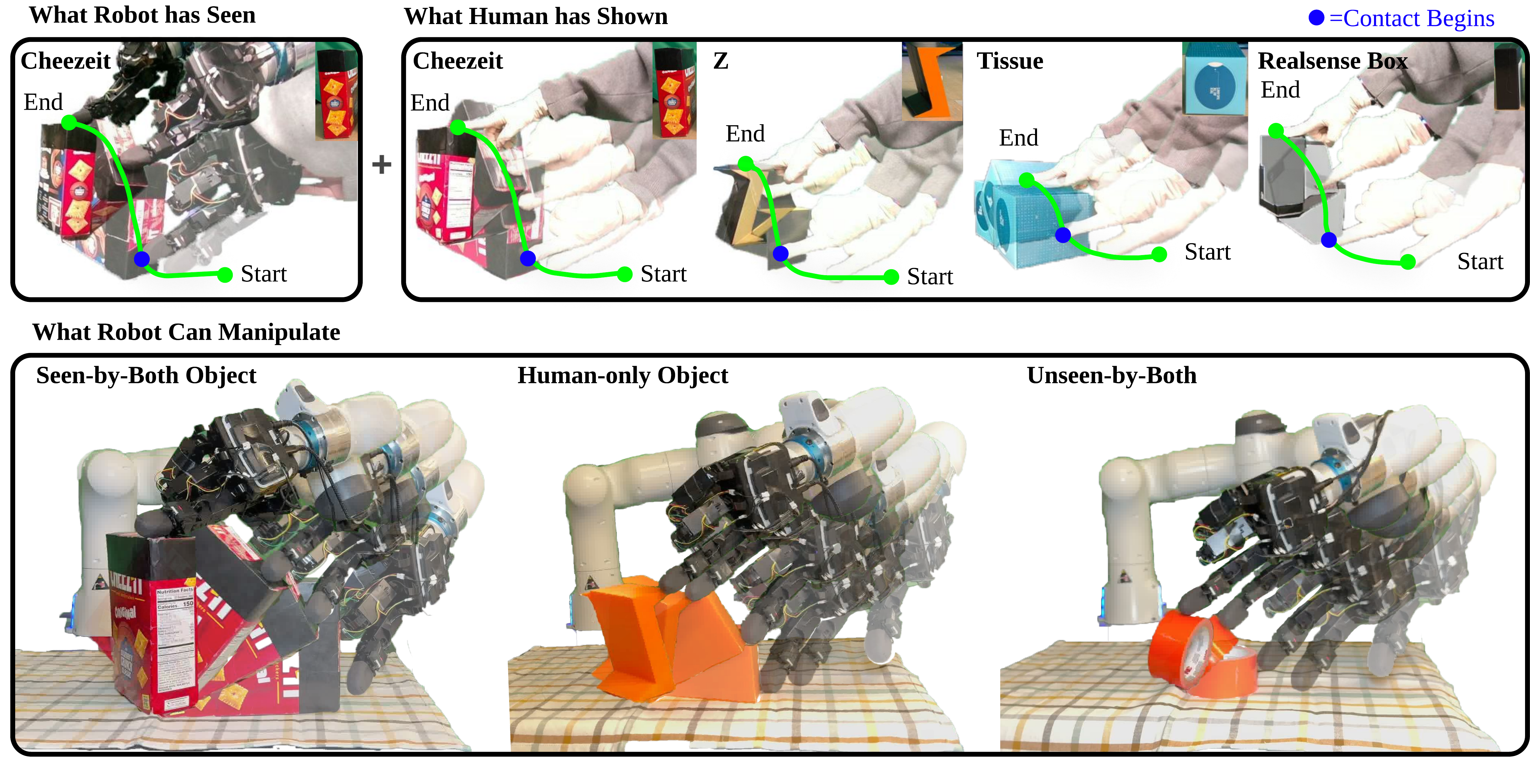}
    \caption{\small \textbf{Pivoting Task.} The task begins in a non-contact state and transitions to pivoting upon contact detection via tactile feedback, with the goal of maintaining contact without dropping the object. \textbf{Top}: training demonstrations. \textbf{Bottom}: robot policy rollouts.}
    \label{fig: result-pivoting}
\end{figure*}

\begin{table*}[t]
\centering
\resizebox{2\columnwidth}{!}{
\begin{tabular}{l c ccc ccc ccc}
\toprule
 & \textbf{} 
 & \multicolumn{6}{c}{\textbf{Seen for Alignment ($v_\theta$, Seen-by-both only)}} 
 & \multicolumn{3}{c}{\textbf{Unseen for Alignment ($v_\theta$)}} \\
 \textbf{Task}
 & \textbf{All}
 & \multicolumn{3}{c}{\textbf{Pivoting}} 
 & \multicolumn{3}{c}{\textbf{Insertion}} 
 & \multicolumn{3}{c}{\textbf{Lid Closing}} \\
 \cmidrule(lr){1-1}
\cmidrule(lr){2-2} 
\cmidrule(lr){3-5} 
\cmidrule(lr){6-8} 
\cmidrule(lr){9-11}
\textbf{Method [\%] $\uparrow$} 
& Avg.$\uparrow$
& Seen-by-both & Human-only & Unseen-by-both 
& Seen-by-both & Human-only & Unseen-by-both  
& Seen-by-both & Human-only & Unseen-by-both  \\
\midrule
Robot Only        
& 38
& 100 & 0 & 0 
& 70 & 0 & 33 
& 100 & 35 & 0 \\

\accro\ w/o Tactile   
& 21
& 0 & 0 & 0 
& 50 & 35 & 43 
& 20 & 0 & 40 \\

\accro\ w/o Align     
& 28
& 0 & 63 & 13 
& 10 & 5 & 0 
& 70 & 40 & 55 \\

\textbf{\accro}      
& \textbf{79}
& \textbf{100} & \textbf{83} & \textbf{60} 
& \textbf{100} & \textbf{65} & \textbf{67} 
& \textbf{100} & \textbf{65} & \textbf{70} \\
\bottomrule
\end{tabular}
}
\caption{\small \textbf{H2R Co-training.} We evaluate human-to-robot transfer on tasks used for alignment (pivoting and insertion) as well as on an unseen task (lid closing). Performance is reported across three object categories: seen-by-both, human-only, and held-out objects. Incorporating human data enables zero-shot generalization to objects not observed during robot training and beyond those present in the human dataset. We compare against three baselines: robot-only training, \accro\ without tactile input, and \accro\ without tactile alignment.}
\label{tab:h2r_results_full}
\end{table*}

\begin{table}[t]
\centering
\setlength{\tabcolsep}{4pt}
\renewcommand{\arraystretch}{0.95}
\small

\resizebox{\columnwidth}{!}{%
\begin{tabular}{cl c ccccccc}
\toprule
& \textbf{Pivoting} & 
& \textbf{Seen-by-both}
& \multicolumn{3}{c}{\textbf{Human-only}}
& \multicolumn{3}{c}{\textbf{Unseen-by-both}} \\
\cmidrule(lr){4-4}
\cmidrule(lr){5-7}
\cmidrule(lr){8-10}
& \textbf{Object}
& \textbf{Avg.$\uparrow$}
& \textbf{Cheezit}
& \textbf{Z}
& \textbf{Tissue}
& \textbf{Realsense}
& \textbf{Tape}
& \textbf{Spray}
& \textbf{CAN Box} \\
\midrule
\multirow{2}{*}{\raisebox{0pt}[7mm][18mm]{\rotatebox{90}{\textbf{Specs}}}}
& Length [mm]      
& 149 & 210 & 140 & 130 & 146 & 100 & 200 & 120 \\
& Weight [g]       
& 224 & 558 & 117 & 191 & 143 & 146 & 366 & 45 \\
\midrule
\multirow{4}{*}{\rotatebox{90}{\textbf{Methods$\uparrow$}}}
& Robot Only        
& 14 & 100 & 0 & 0 & 0 & 0 & 0 & 0 \\
& Ours w/o Tactile   
& 0  & 0 & 0 & 0 & 0 & 0 & 0 & 0 \\
& Ours w/o Align     
& 34 & 0 & 80 & 80 & 30 & 40 & 10 & 0 \\
& \textbf{Ours}      
& \textbf{76} & \textbf{100} & \textbf{100} & \textbf{90} & \textbf{60} & \textbf{30} & \textbf{70} & \textbf{80} \\
\bottomrule
\end{tabular}
}

\vspace{6pt}

\resizebox{\columnwidth}{!}{%
\begin{tabular}{cl c cccccc}
\toprule
& \textbf{Insertion} & 
& \textbf{Seen-by-both}
& \multicolumn{2}{c}{\textbf{Human-only}}
& \multicolumn{3}{c}{\textbf{Unseen-by-both}} \\
\cmidrule(lr){4-4}
\cmidrule(lr){5-6}
\cmidrule(lr){7-9}
& \textbf{Object}
& \textbf{Avg.$\uparrow$}
& \textbf{RVP+}
& \textbf{Mac}
& \textbf{Cana}
& \textbf{Belkin}
& \textbf{TKDY}
& \textbf{FNRSi} \\
\midrule
\multirow{2}{*}{\raisebox{0pt}[7mm][18mm]{\rotatebox{90}{\textbf{Specs}}}}
& Height [mm]       
& 64 & 66 & 80 & 50 & 66 & 50 & 70 \\
& Weight [g]        
& 114 & 83 & 165 & 90 & 80 & 88 & 176 \\
\midrule
\multirow{4}{*}{\rotatebox{90}{\textbf{Methods$\uparrow$}}}
& Robot Only        
& 28 & 70 & 0 & 0 & 50 & 10 & 40 \\
& Ours w/o Tactile   
& 42 & 50 & 20 & 50 & 40 & 50 & 40 \\
& Ours w/o Align     
& 12 & 0 & 10 & 0 & 10 & 50 & 0 \\
& \textbf{Ours}      
& \textbf{72} & \textbf{100} & \textbf{60} & \textbf{70} & \textbf{60} & \textbf{50} & \textbf{90} \\
\bottomrule
\end{tabular}
}

\vspace{6pt}

\resizebox{\columnwidth}{!}{%
\begin{tabular}{cl c ccccc}
\toprule
& \textbf{Lid Closing} & 
& \textbf{Seen-by-both}
& \multicolumn{2}{c}{\textbf{Human-only}}
& \multicolumn{2}{c}{\textbf{Unseen-by-both}} \\
\cmidrule(lr){4-4}
\cmidrule(lr){5-6}
\cmidrule(lr){7-8}
& \textbf{Object}
& \textbf{Avg.$\uparrow$}
& \textbf{Haweek}
& \textbf{Thermos}
& \textbf{Cyxw}
& \textbf{Energify}
& \textbf{Haers} \\
\midrule
\multirow{2}{*}{\raisebox{0pt}[7mm][18mm]{\rotatebox{90}{\textbf{Specs}}}}
& Height [mm]       
& 115 & 95 & 100 & 140 & 130 & 110 \\
& Width [mm]        
& 95 & 100 & 85 & 95 & 93 & 100 \\
\midrule
\multirow{4}{*}{\rotatebox{90}{\textbf{Methods$\uparrow$}}}
& Robot Only        
& 34 & 100 & 70 & 0 & 0 & 0 \\ 
& Ours w/o Tactile   
& 20 & 20 & 0 & 0 & 40 & 40 \\
& Ours w/o Align     
& 52 & 70 & 70 & 10 & 60 & 50 \\
& \textbf{Ours}               
& \textbf{74} & \textbf{100} & \textbf{90} & \textbf{40} & \textbf{70} & \textbf{70} \\
\bottomrule
\end{tabular}
}

\vspace{-4pt}
\caption{\small \textbf{H2R Co-training Results.} Success rates (\%) averaged over 10 rollouts per object for pivoting, insertion, and lid closing. \textbf{Avg.} reports mean performance across all objects for each task. Each object is shown in Sec.~\ref{sec: co-training objects}.}
\label{tab: h2r_results_combined}
\end{table}

\subsection{Human-Robot Policy Co-Training}

\begin{figure*}
    \centering
    \includegraphics[width=0.83\linewidth]{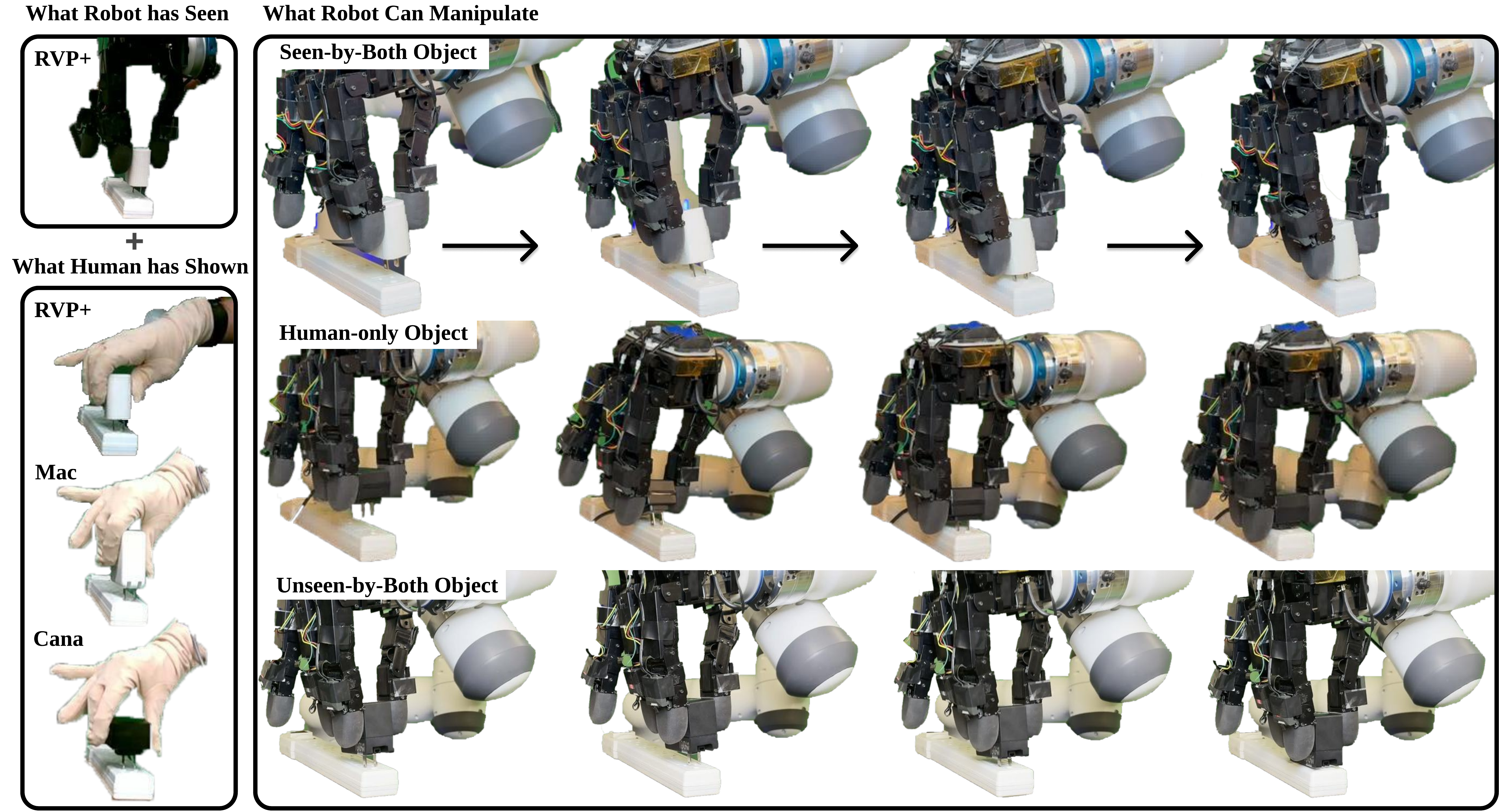}
    \caption{ \small \textbf{Insertion Task.} With randomized grasps, the policy leverages tactile feedback to perform search, alignment, and insertion of the adapter into the outlet. We show human demonstrations helps generalization to unseen adapters across variations in geometry, size, and mass. }
    \label{fig: result-insertion}
\end{figure*}

\begin{figure*}
    \centering
    \includegraphics[width=0.85\linewidth]{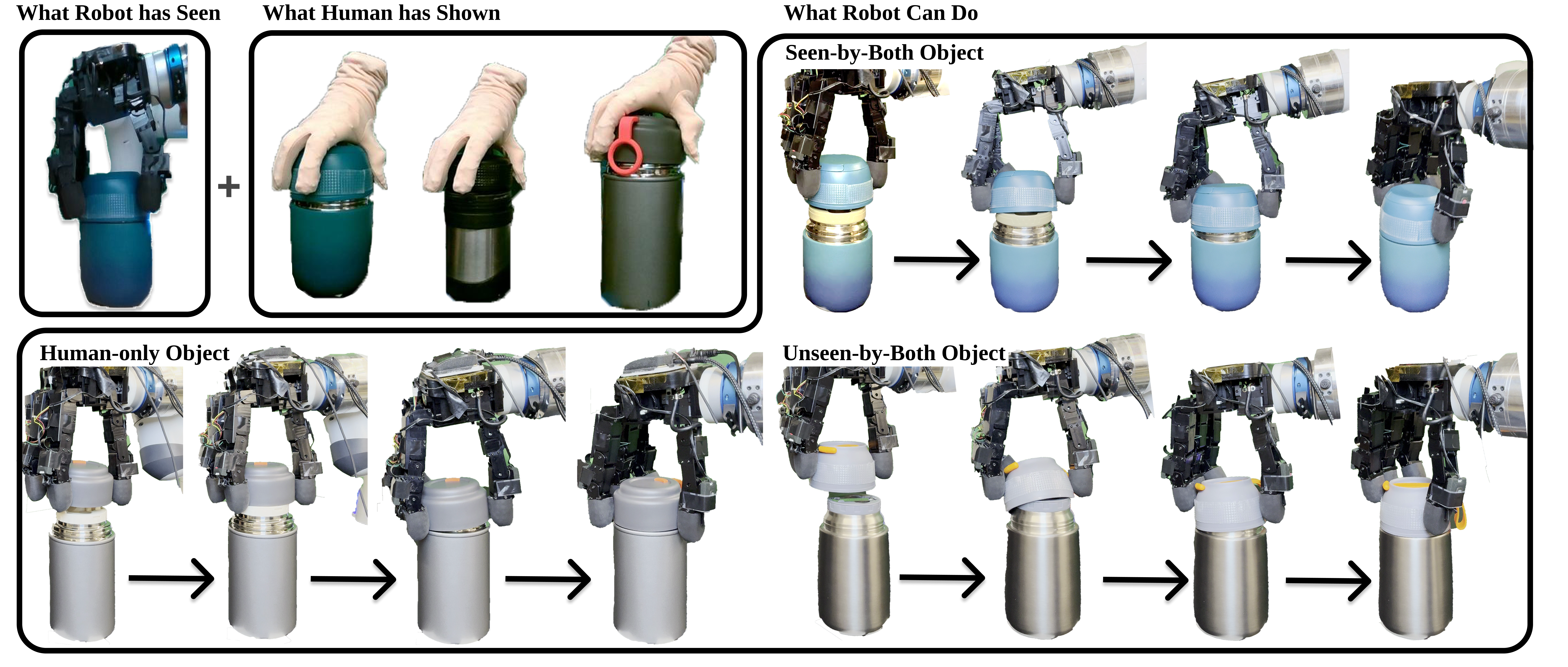}
    \caption{\small\textbf{Lid Closing Task.} With randomized grasps, the policy uses touch to perform search, alignment, and closing between the lid and the bottle.  We show human data improves generalization to unseen objects with varying lid and bottle geometries, sizes, and masses. }
    \label{fig: result-closing}
\end{figure*}

We evaluate the effectiveness of \accro\ for H2R policy co-training on three representative contact-rich manipulation tasks in Tab.~\ref{tab: h2r_results_combined}. All three tasks require force reasoning and begin from a non-contact state, either between the fingertip and the object or between a randomly grasped object and the environment. Successful execution therefore depends on detecting contact onset and reasoning about contact throughout the task as in Fig.~\ref{fig: result-pivoting}, ~\ref{fig: result-insertion}, and ~\ref{fig: result-closing}. The pivoting and insertion tasks evaluate generalization to unseen objects within the same task, while the lid closing task additionally tests our alignment module's generalization to an unseen task class not used during training. We report the co-training results across all tasks in Tab.~\ref{tab:h2r_results_full}. For all tasks, we roll-out 10 times per each object and compare four settings: 

\subsubsection{\textbf{Robot-only baseline}} The robot-only baseline in Tab.~\ref{tab:h2r_results_full} evaluates a policy trained exclusively on robot-collected data using a seen-by-both object. This baseline isolates the contribution of human demonstrations to policy generalization. We observe that augmenting robot training with human demonstrations, which are substantially easier to collect and $\approx4\times$ faster in our setting, markedly improves generalization to unseen objects. Without tactile modality, performance improves +10\% on seen-by-both objects, +59.3\% on human-only objects, and +54.4\% on held-out objects. 


\subsubsection{\textbf{Without Tactile baseline}} The \accro\ w/o tactile baseline in Tab.~\ref{tab:h2r_results_full} evaluates an H2R policy trained without tactile input, relying only on proprioception. This baseline probes the extent to which each task depends on tactile feedback and highlights the role of touch in effective H2R learning. We observe that incorporating tactile input substantially improves performance across all three tasks, increasing success rates +59\% on average. The largest performance gap appears in the pivoting task, where success improves by up to +100\%, followed by lid-closing and insertion. These trends suggest that tasks involving more diverse hand-object interactions, as reflected in the action space, benefit more from tactile feedback, since reliance on proprioception alone provides insufficient information for successful task execution. 

\subsubsection{\textbf{Without Alignment baseline}} The \accro\ w/o align baseline in Tab.~\ref{tab:h2r_results_full} evaluates an H2R policy trained with \accro\ but without tactile alignment, using raw tactile features $h_i$ instead of aligned features $\hat{h}_i$. This baseline isolates the contribution of tactile alignment in H2R co-training. Removing tactile alignment results in a substantial degradation in performance, with an overall $-51\%$ drop in average success rate compared to the full method. Interestingly, non-aligned tactile features are often detrimental, leading to near-complete failure on seen-by-both objects for the pivoting and insertion tasks. For the lid-closing task, which is not used during alignment training, the performance gap between \accro\ and \accro\ w/o align is smaller ($-23\%$ drop). We hypothesize that lid closing admits a broader set of successful contact strategies and can benefit from coarse contact cues present in raw tactile signals, even without consistent cross-embodiment semantics. Together, these results suggest that without alignment, raw tactile features can introduce inconsistent cross-embodiment semantics that hinder policy learning, particularly for tasks requiring precise contact interactions.

\subsubsection{\textbf{\accro}} The \accro\ results in Tab.~\ref{tab:h2r_results_full} correspond to our full method. Across all three object categories, \accro\ demonstrates consistent performance on both tasks used for alignment (pivoting, insertion) and a task not used during alignment or encoder training (lid closing). Specifically, \accro\ achieves success rates of 76\%, 72\%, and 74\% on the pivoting, insertion, and lid-closing tasks, respectively. It also attains perfect performance on seen-by-both objects (100\%) and maintains strong generalization to human-only (71\%) and held-out objects (65.5\%), averaged across three tasks. These results highlight the benefit of tactile alignment combined with data diversity for robust cross-task and cross-object generalization.

\begin{table}
\centering
\resizebox{\columnwidth}{!}{%
\setlength{\tabcolsep}{6pt}
\begin{tabular}{l c c c}
\toprule
\textbf{Light Bulb Screwing} & \textbf{Ours w/o Tactile} & \textbf{Ours w/o Align} & \textbf{Ours} \\
\midrule
Success rate$\uparrow$[\%]  & 0 & 0 & 100 \\
\bottomrule
\end{tabular}
}
\caption{\small \textbf{Light Bulb Screwing Result.} Success rates on the light bulb screwing task, evaluated over 10 rollouts.}
\label{tab:lightbulb}
\end{table}

\subsection{Dexterous Robot Policy Learning with Human Data Only}

\begin{figure}[h]
    \centering
    \includegraphics[width=0.8\linewidth]{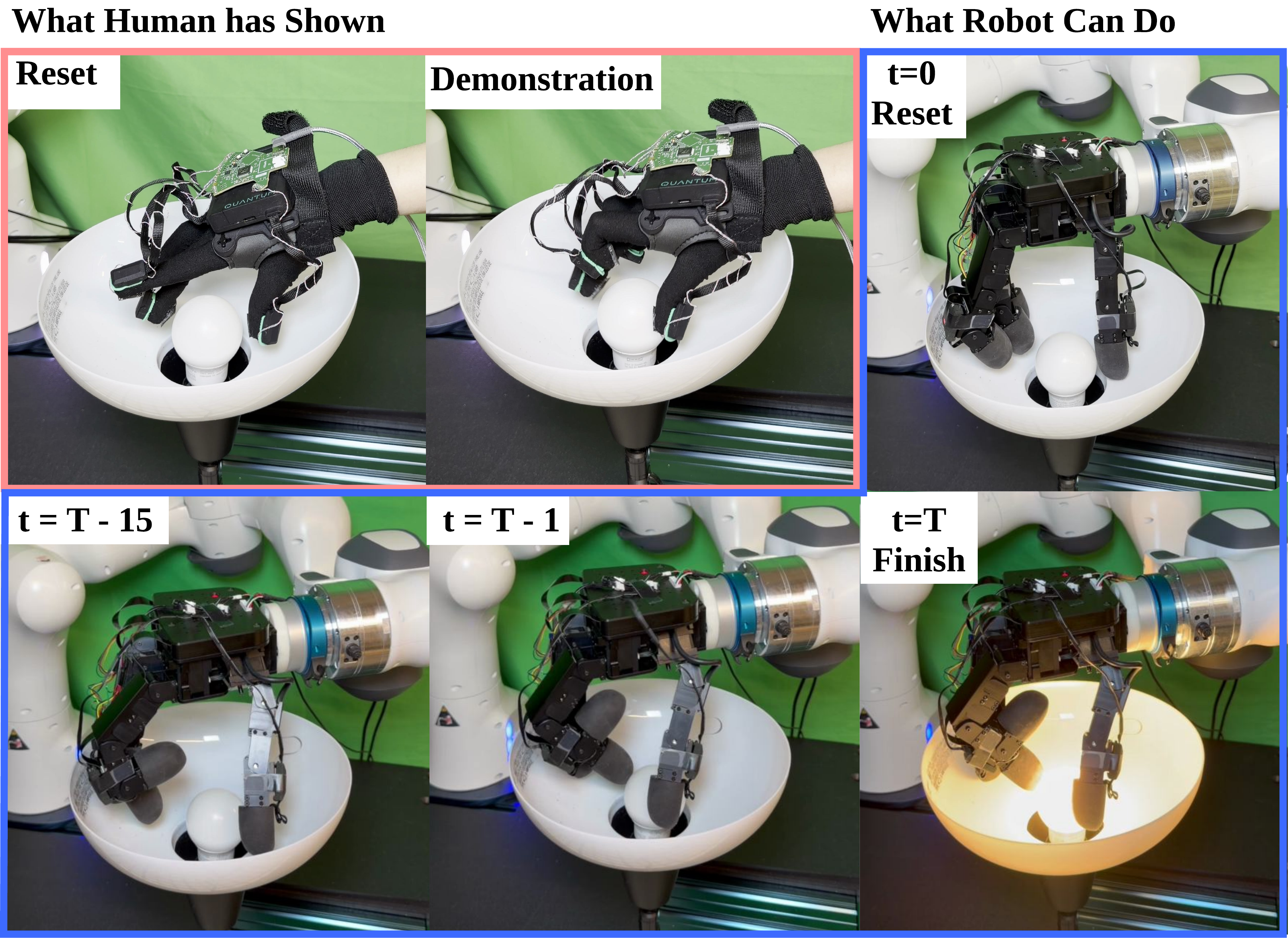}
    \caption{\small \textbf{Light Bulb Screwing Task.} \accro\ learns a dexterous tactile policy using only human demonstrations, with zero robot data.}
    \label{fig: light bulb screwing}
\end{figure}

The light-bulb screwing task in Fig.~\ref{fig: light bulb screwing} represents a dexterous and occlusion-heavy manipulation scenario, often due to the presence of lamp shades and small objects compared to robot fingers. As a result, teleoperation is non-trivial for this dexterous task \cite{yin2025dexteritygen}, particularly in the absence of tactile feedback and reliable visual cues. In this task, we highlight two aspects of human-to-robot transfer: (1) the dexterity enabled by human demonstrations, in which demonstrators rely on tactile feedback to guide precise finger motions during screwing, and (2) zero-shot human-to-robot policy execution. As in Tab.~\ref{tab:lightbulb}, our \accro\ policy achieves a 100\% success rate. On average, screwing the light bulb until illumination takes approximately 61 seconds. In contrast, without tactile input, the success rate drops to 0\%. The primary failure mode is the inability of the fingertips to establish stable contact with the light bulb, preventing task execution altogether. Without alignment, the success rate is also 0\%, with failures primarily arising from jamming, from which the policy cannot recover, often leading to complete unscrewing of the light bulb. Compared to H2R co-training (Tab.~\ref{tab:h2r_results_full}), the performance gap between the two baselines is largest, highlighting the importance of tactile alignment, especially when robot data are scarce.


\subsection{H2R Force Estimation \label{sec: force estimation}}

We quantitatively evaluate H2R tactile alignment using a cross-sensor force prediction task, where neither tactile observations nor force measurements are seen during encoder or alignment training. This evaluates whether the aligned latent space preserves physically meaningful information that transfers across heterogeneous tactile sensors.

\subsubsection{\textbf{Evaluation}} We train a robot force decoder $D^r(r_j) \in \mathbb{R}^3$ that predicts contact forces from robot tactile features. $D^r$ is trained on top of frozen  \accro\, analogous to a linear probe as in~\cite{assran2023self} (Appendix~\ref{sec: force decoder architecture}). We evaluate force prediction under three settings:
\emph{(i)} $H \rightarrow R$ \textit{without} alignment, where human tactile features are directly decoded as $D^r(h_i)$;
\emph{(ii)} $H \rightarrow R$ with \accro, where the human tactile features are first mapped into the aligned latent space and decoded as $D^r(g_\theta(h_i))$; and
\emph{(iii)} $R \rightarrow R$, where the evaluation are performed on heldout robot data, representing a best-case upper bound. We evaluate five times and report the the mean and standard deviation of the per-axis $\ell_1$ force prediction error.

\subsubsection{\textbf{Analysis}} 
\begin{figure}
    \centering
    \includegraphics[width=0.75\linewidth]{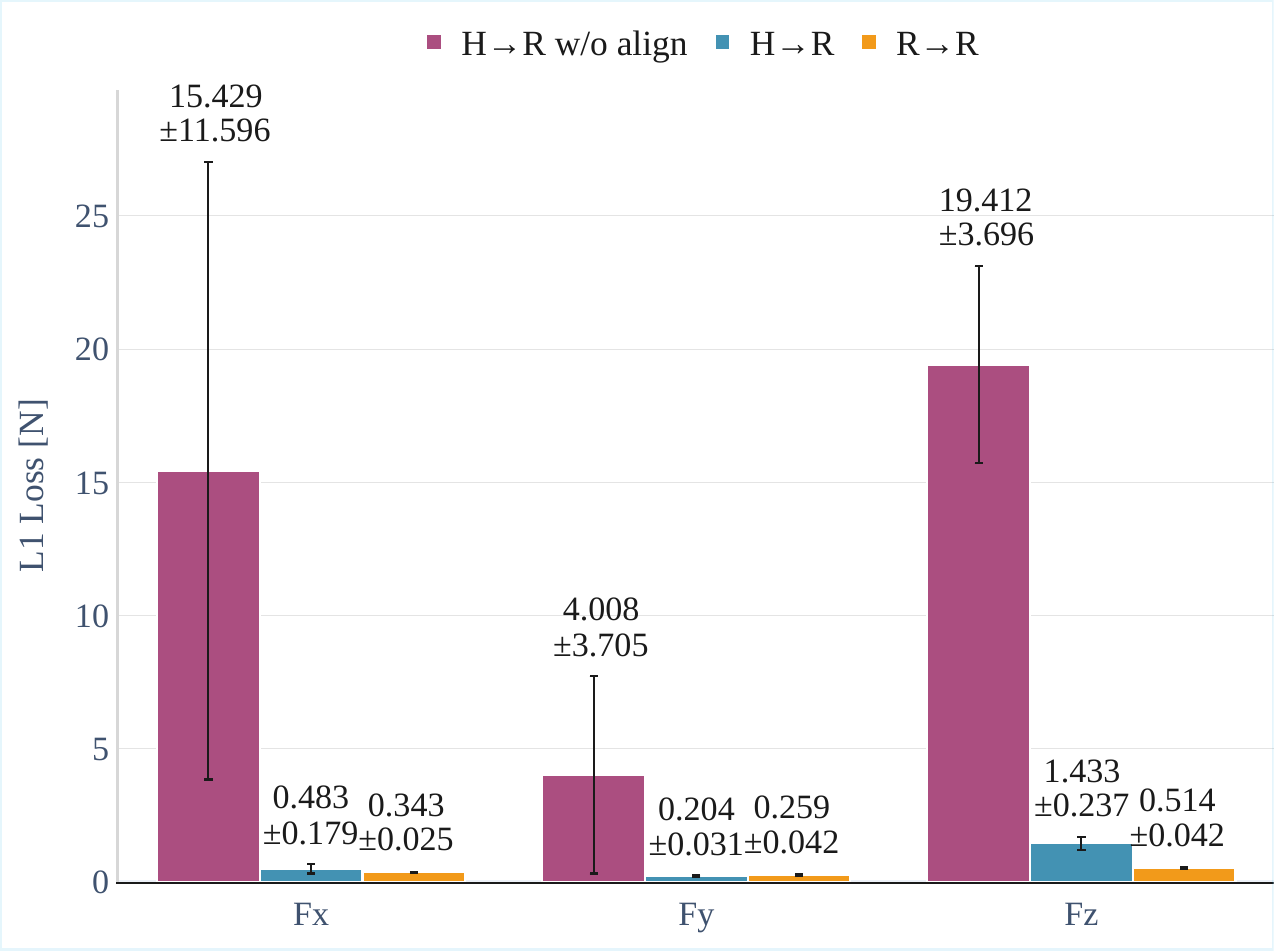}
    \caption{ \small $\ell_1$ force prediction error (mean $\pm$ std) along each axis, averaged over five evaluations.
$G \rightarrow R$ evaluates force prediction on the robot using human tactile signals, with (blue) and without alignment (red) alignment. With alignment reduces the force prediction error by 96.75\% across the three force axes. $R \rightarrow R$ denotes the robot-to-robot baseline (yellow), representing a best-case scenario.
}
\label{fig: force prediction}
\end{figure}

Fig.~\ref{fig: force prediction} reports the $\ell_1$ force prediction error along each axis. In the $H \rightarrow R$ setting without alignment, force prediction errors are high across all axes, while \accro\ reduces the force prediction error by approximately $98\%$, $99\%$, and $93\%$ along the $F_x$, $F_y$, and $F_z$ axes, respectively, while also significantly reducing variance across runs. One reason for the large error without alignment is that the force decoder $D^r$, trained only on robot tactile features, must extrapolate across the substantial distribution gap. To contextualize, we compare against the robot-to-robot ($R \rightarrow R$) baseline, which represents a best-case upper bound where both training and evaluation are performed on robot tactile data. With alignment, $H \rightarrow R$ performance reaches within $2\%$ of the $R \rightarrow R$ baseline on $F_x$ and within $13\%$ on $F_y$. A larger gap remains on $F_z$, where the aligned error is on average $0.7\mathrm{N}$ higher than the $R \rightarrow R$ baseline. Nevertheless, alignment recovers a large portion of the performance gap, implying \accro's ability to transfer physically meaningful force information between heterogeneous tactile sensors. 


\section{Limitation}
While the formulation of \accro\ is sensor-agnostic, our evaluation is limited to a single glove-robot pairing due to hardware availability. Extending the approach to additional tactile modalities (e.g., vision-based tactile sensors), multi-hand settings, or full-palm sensing is left for future work. Moreover, tactile alignment alone does not address visual discrepancies between human and robot embodiments. Incorporating vision and other modalities into a unified multi-modal policy is also an important direction for future work. 



\clearpage
\section*{Acknowledgments}
We thank Chan Hee Song, Mark van der Merwe, Tingfan Wu, Taosha Fan, and François Hogan for valuable project discussions. We are grateful to Mike Lambeta, James Lorenz, and Fan Yang for their support with hardware debugging, and to Woniks for their Allegro Hand hardware support.

\bibliographystyle{plainnat}
\bibliography{references}

\clearpage
\begin{appendices}

\section{Dataset}
\subsection{Wearable Devices \label{sec: wearable devices}}
\begin{figure}[h]
    \centering
    \includegraphics[width=0.7\linewidth]{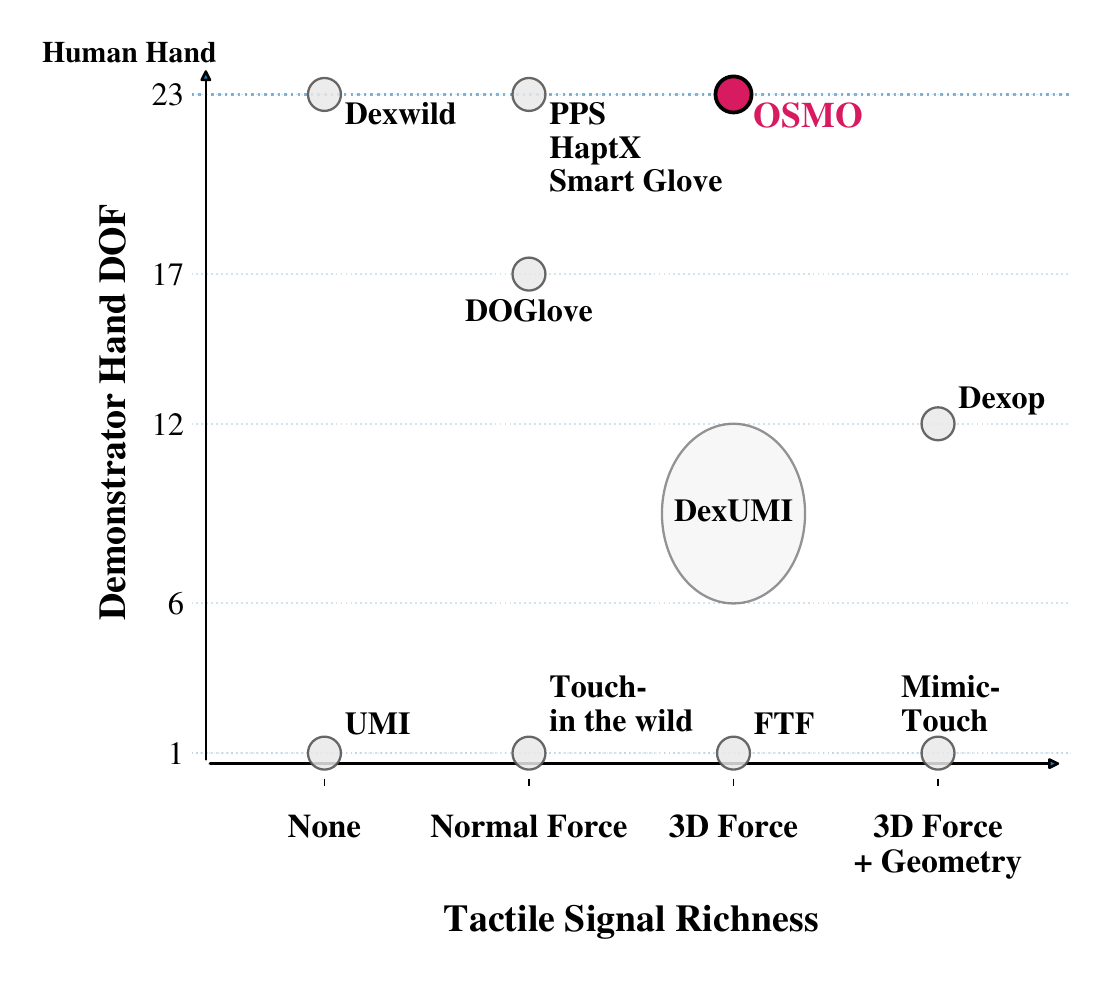}
    \caption{Our human dataset leverages OSMO \cite{yin2025osmo} tactile glove, providing the human demonstrator with full dexterity while capturing both shear and normal tactile signals.}
    \label{fig: hardware comparison}
\end{figure}

The human hand is a remarkably dexterous manipulator, with up to 23 degrees of freedom~\cite{yang2025lightweight}. In contrast, robotic manipulators are undergoing rapid evolution: recent platforms exhibit increasing degrees of freedom alongside a growing diversity of mechanical designs and actuations ~\cite{sharpa_hand, wuji_hand, clone_hand}. The combination of fast hardware advancement and heterogeneous robot embodiments poses a challenge for data-driven policy learning, as demonstrations collected on any single platform risk becoming quickly outdated or narrowly tailored to a specific design. This motivates the need for a more generalizable source of dexterous data that is not tied to a particular robot. In this work, we therefore propose to treat the human hand as a universal embodiment for data collection, viewing human demonstrations as a shared data source that could, in principle, be leveraged across a wide range of robot hands as in \cite{tao2025dexwild, yin2025osmo}. This approach differs from other hardware interfaces such as exoskeleton gloves (e.g., DOGLove \cite{zhang2025doglove}, Dexop \cite{fang2025dexop}, DexUMI \cite{xudexumi} in Fig.~\ref{fig: hardware comparison}), which constrain natural hand motion and tend to produce data that is specific to a single robotic embodiment.

\subsection{Policy Dataset}
For both the pivoting and insertion tasks, we reuse the `seen-to-both' object demonstrations from the cross-sensor tactile alignment dataset.

\subsection{Force Prediction Data-collection Setup \label{sec: force data collection setup}}
\begin{figure}[h!]
    \centering
    \includegraphics[width=1\linewidth]{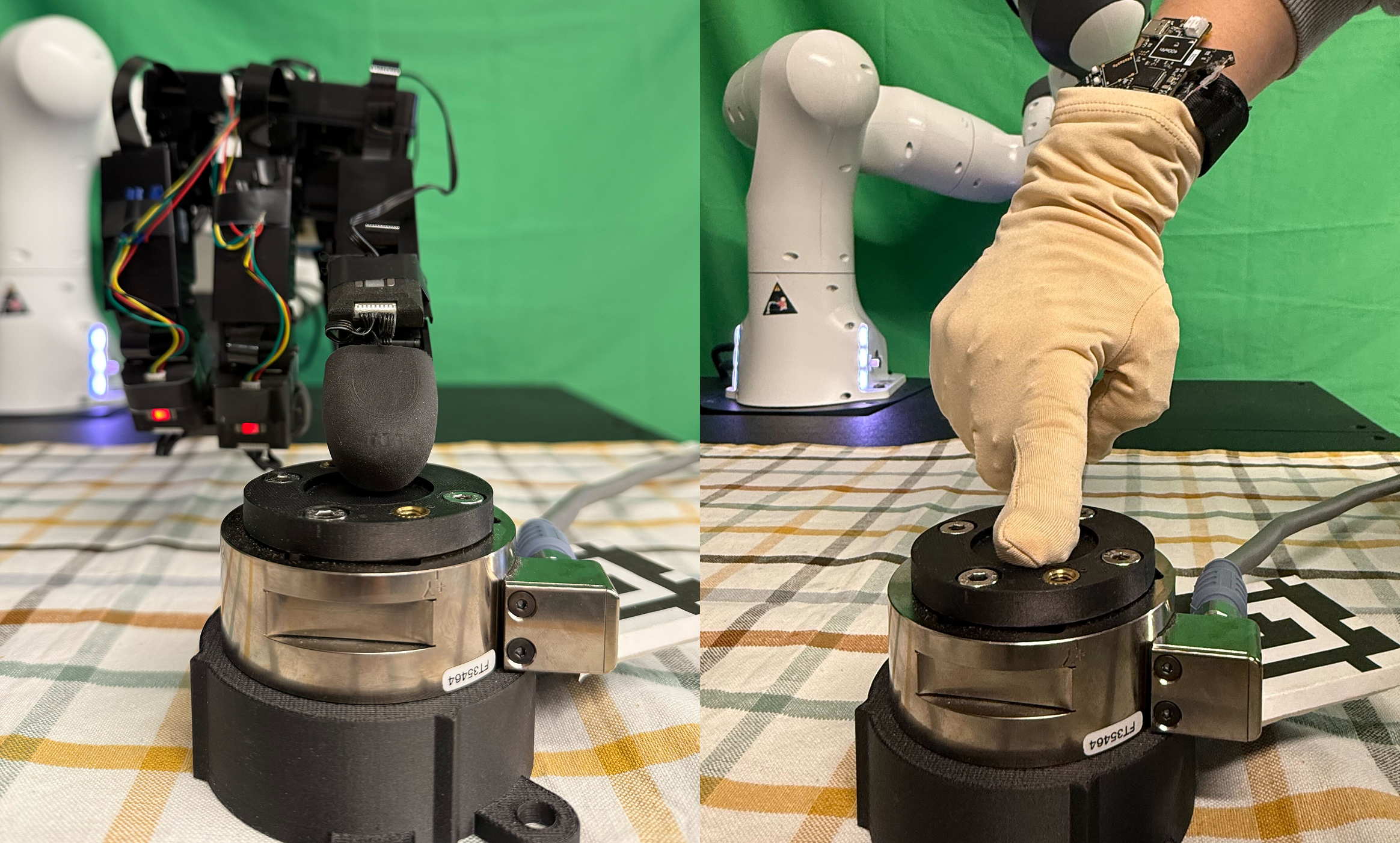}
    \caption{Robot and human force label collection using an F/T sensor (ATI Gamma) fixed beneath the table.}
    \label{fig: force data collection setup}
\end{figure}

Figure~\ref{fig: force data collection setup} illustrates our force data collection setup. An ATI Gamma force-torque sensor is rigidly mounted to the tabletop to provide a stable reference frame for all interactions. During data collection, we align both the Xela tactile sensor and the glove fingertip along the +x direction of the sensor frame to ensure consistent contact orientation across trials.

\subsection{Force Prediction Dataset}

\begin{figure*}
    \centering
    \includegraphics[width=1\linewidth]{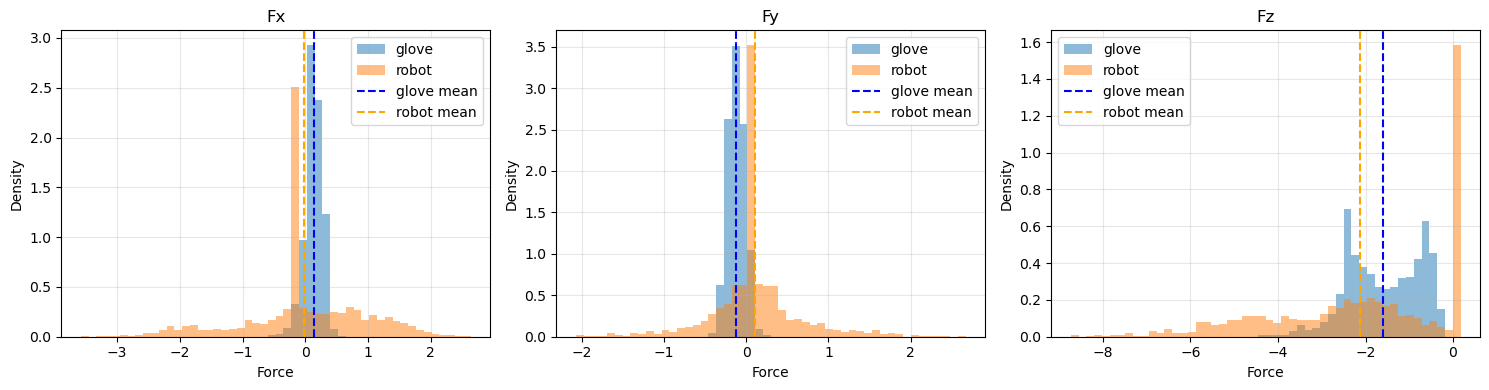}
    \caption{Force dataset distribution}
    \label{fig: force dataset distribution}
\end{figure*}
The robot dataset’s train and test sets were randomly sampled from data whose ground-truth forces are approximately centered around the mean of (–0.0185, 0.1003, –2.1348) in the (x, y, z) directions. Similarly, the glove dataset’s train and test sets were randomly sampled from forces centered around the mean of (0.1347, –0.1228, –1.5903). Overall, the mean x- and y-forces of the glove and robot datasets are comparable, whereas the robot dataset exhibits an approximately 0.5 N larger mean force in the z-direction.

\begin{figure}[h!]
    \centering
    \includegraphics[width=1\linewidth]{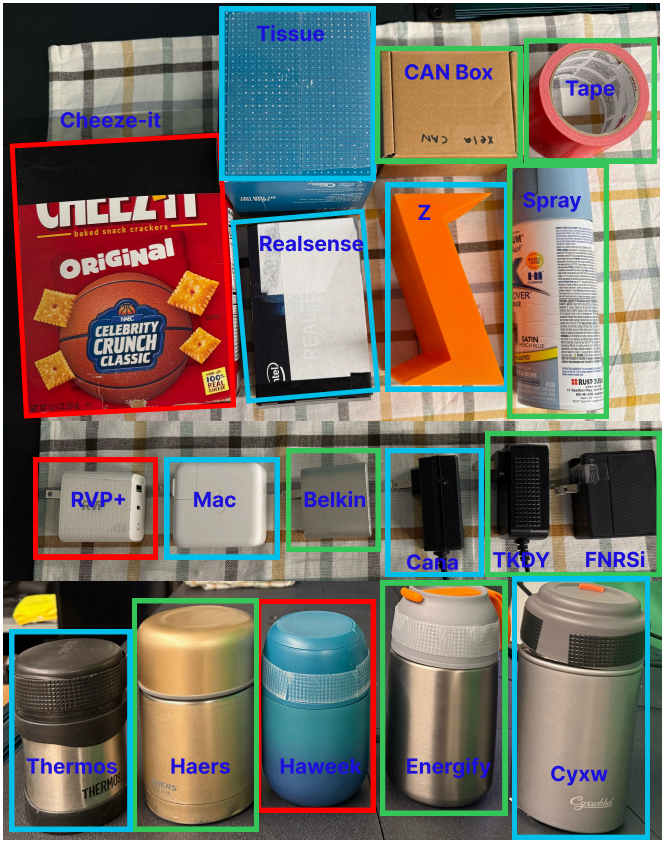}
    \caption{Red boxes denote `Seen-by-all' objects, blue boxes denote `Human-only' objects, and green boxes denote `Held-out' objects for each task in Tab.~\ref{tab: h2r_results_combined}.}
    \label{fig: cotraining objects}
\end{figure}

\subsection{H2R Policy Co-training Objects \label{sec: co-training objects}}

Fig.~\ref{fig: cotraining objects} illustrates the evaluation objects and their corresponding names used in Tab.~\ref{tab: h2r_results_combined}. We also visualize the object categories using color-coded bounding boxes. As shown in the figure, we intentionally select objects with diverse geometries and physical properties to assess generalization across a broad range of manipulation settings.

\section{Post Processing}

\subsection{Data Post Processing \label{sec: data post processing}}
In this work, we extract fingertip locations and wrist poses using WiLoR \cite{potamias2025wilor}, with hand masks obtained from SAM3 \cite{carion2025sam3segmentconcepts}. To improve hand pose accuracy, we refine the estimated hand meshes using ICP \cite{rusinkiewicz2001efficient} with depth maps from FoundationStereo \cite{wen2025stereo}, and apply a Savitzky–Golay filter \cite{savitzky1964smoothing} for temporal consistency.

For object pose estimation, we propose a mesh-less object pose extraction pipeline adapted from Any6D \cite{lee2025any6d}. We use FoundationPose \cite{foundationposewen2024} for object pose estimation on depth maps from FoundationStereo \cite{wen2025stereo}. To run FoundationPose, we perform automatic mesh reconstruction once per task for both the robot and glove datasets, using the first frame of the first human demonstration $\mathcal{T}^h_1$. Object meshes are automatically generated from RGB images using an object name prompt via SAM3 \cite{carion2025sam3segmentconcepts} and SAM3D \cite{sam3dteam2025sam3d3dfyimages}. SAM3 \cite{carion2025sam3segmentconcepts} segments the object from the RGB image, and SAM3D reconstructs textured object meshes from the segmented object, even under occlusions.

With FoundationPose \cite{foundationposewen2024}, we estimate the object pose directly on the initial frame of each trajectory, and then switch to tracking mode for subsequent frames to ensure temporal consistency. Although the objects are symmetric, SAM’s texture generation helps disambiguate symmetry when used with FoundationPose (e.g., for the Cheez-It box). For objects with symmetric adapter textures, we use the initial object pose from the first human trajectory ($\mathcal{T}^h_1$) to align the object axes, assuming that objects share approximately the same initial state as specified in the problem statement.

Our entire pipeline is executed sequentially in a single pass and without the use of any privileged information, such as pre-existing 3D CAD models. We demonstrate this capability on pivoting with a box (Cheez-It) and insertion with an adapter (RVP+).

\subsection{Signal Post Processing \label{sec: signal post processing}}

\textbf{Human} For the human glove data, we first estimate and remove the baseline by measuring the non-contact state at the beginning and end of each trajectory and subtracting it from the raw signal. The current OSMO~\cite{yin2025osmo} glove exhibits sign flips depending on the underlying magnetic skin quadrant under the same applied force; to avoid ambiguity in contact location, we take the absolute value of the tactile signal. Glove data are collected at 30 fps. 

\textbf{Robot:} Robot tactile data are collected at 100 fps; since the Xela sensor signal drifts over time, we record a baseline before every data collection session or robot rollout and subtract it accordingly.

\section{Pseudo-pairs}
\subsection{Pose Normalization\label{sec: appendix psuedo-pairs}}

For each task $\tau$, we compute normalization statistics from the robot dataset and apply the same normalization to both human and robot data belonging to that task. Specifically, we subtract the mean and divide by the standard deviation, where the standard deviation is taken as the maximum across axes. This normalization is performed separately for position and orientation.

\medskip
\subsubsection{Fingertip position normalization}  

Let
\[
\mathcal{P}_\tau^r = \{ p_{t,k}^r \mid (F_t^r, P_t^r, w_t^r, o_t^r)\in A_\tau^r \}
\]
denote the set of robot fingertip positions for task $\tau$. We compute the mean from the robot data as
\begin{align}
\mu_p(\tau)
= \frac{1}{|\mathcal{P}_\tau^r|}\sum_{p\in \mathcal{P}_\tau^r} p,
\end{align}
and the maximum per-axis standard deviation as
\begin{align}
\sigma_{p,\max}(\tau)
= \max_{d\in\{x,y,z\}}
\operatorname{std}\!\left(\{p_d : p\in \mathcal{P}_\tau^r\}\right).
\end{align}

We then normalize each fingertip position for \emph{both} embodiments as
\[
\tilde p_{t,k}^e
= \frac{p_{t,k}^e - \mu_p(\tau)}{\sigma_{p,\max}(\tau)}, 
\qquad e\in\{h,r\}.
\]

\medskip
\subsubsection{Object pose normalization}  

We write the object pose as $o_t^e = (s_t^e, q_t^e)$, where $s_t^e\in\mathbb{R}^3$ is position and $q_t^e\in\mathbb{R}^3$ is a rotation vector. Define the set of robot object positions for task $\tau$ as
\[
\mathcal{S}_\tau^r = \{ s_t^r \mid (F_t^r,P_t^r,w_t^r,o_t^r)\in A_\tau^r \}.
\]

We compute the mean from the robot data as
\[
\mu_s(\tau) = \frac{1}{|\mathcal{S}_\tau^r|}\sum_{s\in \mathcal{S}_\tau^r} s,
\]
and the maximum per-axis standard deviation as
\[
\sigma_{s,\max}(\tau)
= \max_{d\in\{x,y,z\}}
\operatorname{std}\!\left(\{s_d : s\in \mathcal{S}_\tau^r\}\right).
\]

We normalize object positions for both embodiments as
\[
\tilde s_t^e
= \frac{s_t^e - \mu_s(\tau)}{\sigma_{s,\max}(\tau)}, 
\qquad e\in\{h,r\}.
\]

For orientations, represented as rotation vectors, we apply the same procedure. Define the set of robot orientations for task $\tau$ as
\[
\mathcal{Q}_\tau^r = \{ q_t^r \mid (F_t^r,P_t^r,w_t^r,o_t^r)\in A_\tau^r \}.
\]

We compute the mean from the robot data as
\[
\mu_q(\tau) = \frac{1}{|\mathcal{Q}_\tau^r|}\sum_{q\in \mathcal{Q}_\tau^r} q,
\]
and the maximum per-axis standard deviation as
\[
\sigma_{q,\max}(\tau)
= \max_{d\in\{x,y,z\}}
\operatorname{std}\!\left(\{q_d : q\in \mathcal{Q}_\tau^r\}\right).
\]

We normalize object orientations for both embodiments as
\[
\tilde q_t^e
= \frac{q_t^e - \mu_q(\tau)}{\sigma_{q,\max}(\tau)}, 
\qquad e\in\{h,r\}.
\]

\subsection{Balancing Term \label{sec: balancing term}}
In all experiments, we use a fixed balancing term of $\lambda=1$ across all tasks and embodiments. To assess the sensitivity of our method to this choice, we report the alignment performance under neighboring values of $\lambda$. Table~\ref{tab:lambda_sensitivity} shows the EMD reduction rate before and after alignment for different $\lambda$. We observe that the performance remains stable over a reasonably wide range of $\lambda$, indicating that our method is not sensitive to this hyperparameter.

\begin{table}[h]
    \centering
    \begin{tabular}{cccccc}
        \toprule
        $\lambda$ & 0.8 & 0.9 & 1.0 & 1.1 & 1.2\\
        \midrule
        EMD Red. [\%] $\uparrow$ & 80.5 & 82.5 & 83.2 & 80.6 & 79.2\\
        \bottomrule
    \end{tabular}
    \caption{EMD reduction rate (EMD Red.) before and after alignment for different values of $\lambda$. Higher values indicate better alignment between human and robot tactile distributions. Performance remains robust across a range of $\lambda$.}
    \label{tab:lambda_sensitivity}
\end{table}


\subsection{Further Clarification of Pseudo-Pair Construction}

In Sec.~\ref{sec: pseudo-pair}, we define the pseudo-pair set $P$ using a similarity threshold $\delta$ for notational compactness. In practice, $\delta$ primarily serves as a validity constraint in pseudo-pair construction, as we perform nearest-neighbor matching subject to this threshold. Specifically, for each human finger-object transition $O^h_i$, we select the $N$ nearest robot transitions under the similarity metric $S(O^h_i, O^r_j)$. The threshold $\delta$ is then applied as a lower bound to exclude degenerate matches (e.g., those with large pose discrepancies). Intuitively, if $\delta$ is set too large, it can admit low-quality or semantically meaningless pairs into $P$. However, within a reasonable range, the resulting pseudo-pair set and consequently our alignment performance remains largely consistent.

To validate this, we perform a sensitivity analysis of the tactile alignment with respect to $\delta$, as shown in Tab.~\ref{tab:lambda_sensitivity}. The results indicate that $\delta$ can take a broad range of values without significantly affecting performance. All experiments in the main paper therefore use a fixed $N$ and a single reasonable $\delta$ (with $N=3$ and $\delta=2.0$).

\begin{table}[h]
    \centering
    \begin{tabular}{cccccc}
        \toprule
        $\delta$ & 1.5  & 2.0  & 2.5 & 3.0 & 3.5 \\
        \midrule
        EMD Red. [\%] $\uparrow$ & 82.7 & 83.2& 83.9 & 80.7 & 83.9\\
        \bottomrule
    \end{tabular}
    \caption{EMD reduction rate (EMD Red.) before and after alignment for different values of $\lambda$ for different $\delta$.}
    \label{tab:lambda_sensitivity}
\end{table}

\subsection{Binary Contact Threshold}
\begin{figure}
    \centering
    \includegraphics[width=1\linewidth]{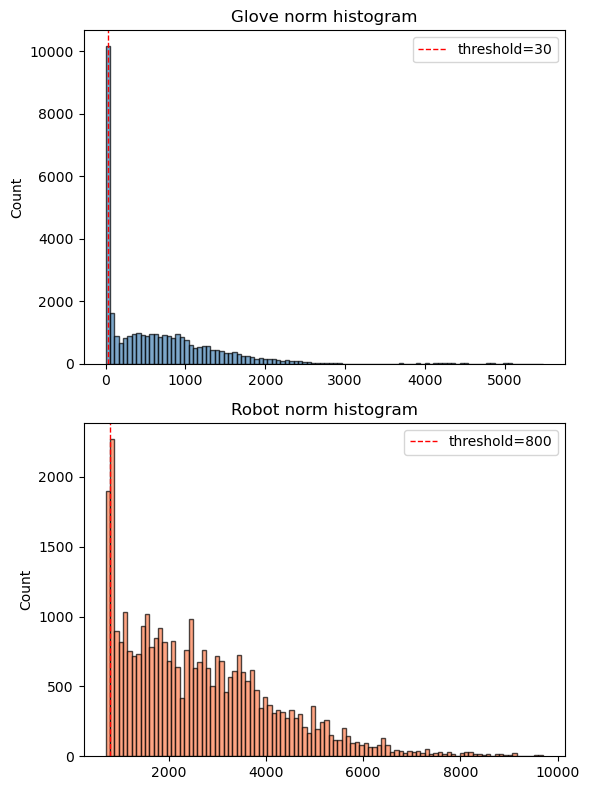}
    \caption{Norm of raw sensor signal observations from the human glove~\cite{yin2025osmo} and the robot Xela sensor. The red dotted vertical line indicates the chosen non-contact threshold.}
    \label{fig: norm hist}
\end{figure}

To motivate our choice of non-contact thresholds, we visualize the distribution of raw tactile signal norms for both human and robot sensors in Fig.~\ref{fig: norm hist}. In practice, non-contact signals tend to exhibit a high-density peak near zero, whereas when contact occurs, the signal magnitude typically changes significantly. This clear separation makes threshold-based detection of contact straightforward. We therefore select thresholds that are small enough to capture the dominant non-contact peak in the histogram while excluding meaningful contact signals. Specifically, we set $\delta_h = 1200$ for the human glove and $\delta_r = 30$ for the robot Xela sensor.

\section{Training and Architecture Details}

\subsection{Alignment Module}
\label{sec: tactile alignment training details}

In our implementation, we parameterize $\textcolor{sharedpurple}{v_\theta}$ using a compact multilayer perceptron with three hidden layers of width 1024. Following~\cite{liuflow}, we train the model with 100 discretized time steps for 200,000 epochs using a learning rate of $5\times 10^{-5}$. In our case, the entire training procedure takes about 10 minutes on a single NVIDIA RTX 4090.

\subsection{H2R Policy -- Pivoting}

For pivoting, the policy takes as input only the index finger’s tactile and proprioceptive signals. We treat the index fingertip as a representative of the many fingers involved in manipulation. The action space is 6-dimensional, corresponding to the index fingertip position and wrist rotation, and both the input and action spaces are defined with respect to the robot base frame.

\subsection{H2R Policy -- Insertion}

For insertion, only three fingers (thumb, index, and middle) are actively engaged. Accordingly, the policy input consists of the tactile and proprioceptive signals from these three fingers. The 6-dimensional action output represents the index fingertip position and wrist rotation (expressed in the robot base frame), as this task leverages only wrist motion.

\subsection{H2R Policy -- Lid Closing}

For lid closing, all four fingers (thumb, index, middle, and ring) are engaged. Thus, the policy input includes tactile and proprioceptive signals from all four fingertips. Similar to insertion, this task also leverages only wrist motion, and the 6-dimensional action output corresponds to the index fingertip position and wrist rotation, defined in the robot base frame.

\subsection{Force Decoder}
\label{sec: force decoder architecture}

The force decoder is implemented as a multilayer perceptron with two hidden layers of 64 units each and ReLU activations. It is trained using an $\ell_2$ regression loss with a learning rate of $10^{-4}$ for 100k epochs.

\subsection{H2R Policy -- Lightbulb Screwing}

For lightbulb screwing, the policy uses all four fingertip positions as both input and output, expressed in the robot base frame. In this case, we do not include wrist rotation in the action space, as the wrist remains stationary throughout the interaction.

\subsection{Learned Rectified Flow}
\begin{figure}[h]
    \centering
    \includegraphics[width=1\linewidth]{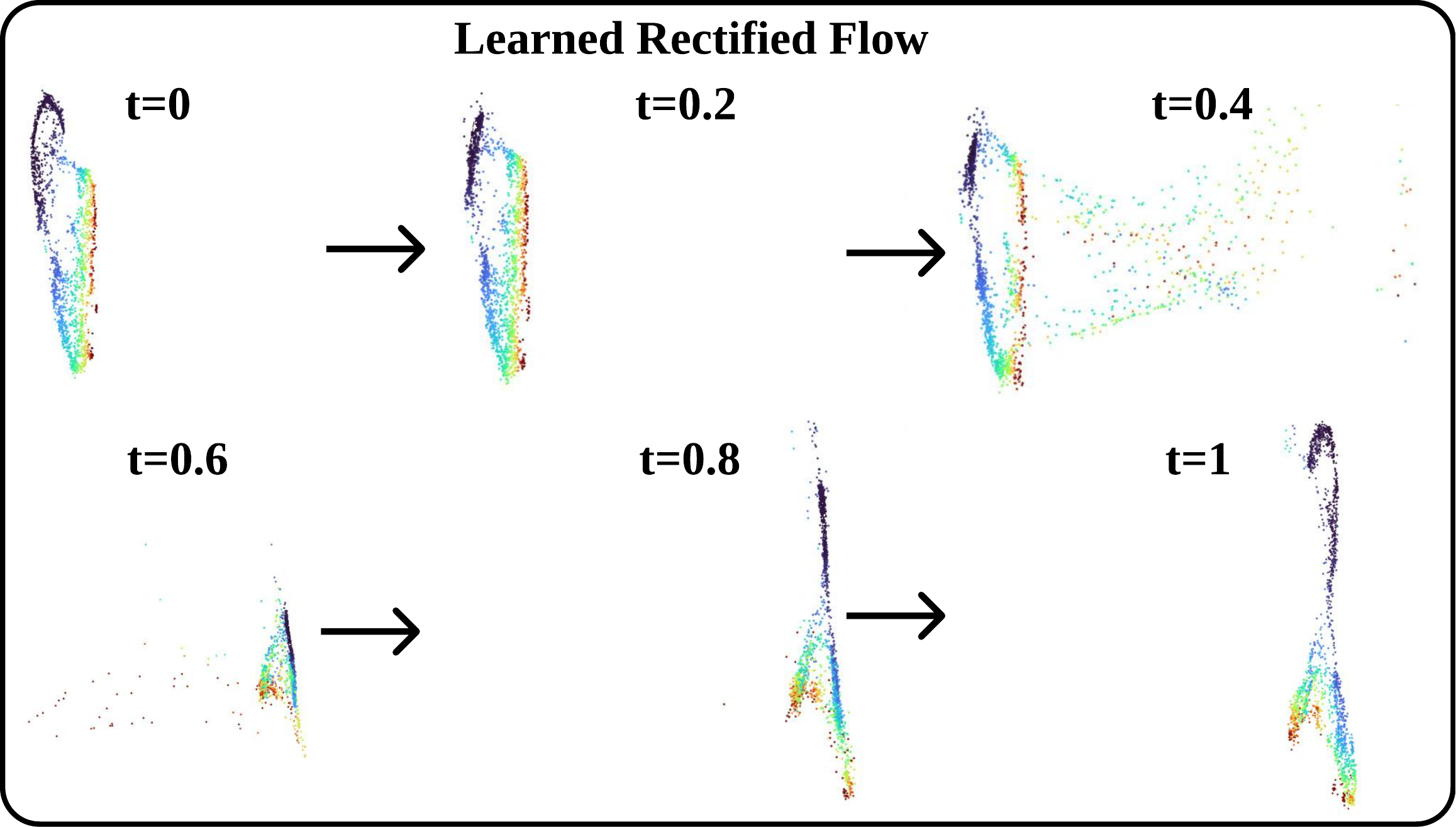}
    \caption{How the tactile latent distribution of human's evolve with time via rectified flow.}
    \label{fig: flow}
\end{figure}

Fig.~\ref{fig: flow} shows how the glove latent evolves to match the robot distribution. This figure shares the same color coding as Fig.~\ref{fig: tactile latent distribution}.

\section{Inference \label{sec: policy roll-out details}}
\subsection{Policy Rollouts}

During inference, the policy operates at 10 Hz with an action chunk size of 32. For pivoting, we execute 4 actions per rollout; for insertion, we execute 2 actions; and for lid closing, we execute 8 actions. During rollouts, the predicted action representations are converted into a sequence of joint commands via inverse kinematics, solved using PyRoki~\cite{kim2025pyroki}. For all co-training tasks with wrist-only actions, we convert the predicted index-fingertip location and wrist orientation into a wrist pose in $\mathrm{SE}(3)$ by assuming a rigid transformation between the wrist and the index fingertip.

For light-bulb screwing, the policy runs at a higher frequency of 30 Hz, and we execute 12 actions from the 32-step action chunk to enable fine-grained manipulation at a higher control rate.
 \end{appendices}
\end{document}